\documentclass{article}

\usepackage{PRIMEarxiv}

\usepackage[utf8]{inputenc} 
\usepackage[T1]{fontenc}    
\usepackage{hyperref}       
\usepackage{url}            
\usepackage{booktabs}       
\usepackage{amsfonts}       
\usepackage{nicefrac}       
\usepackage{microtype}      
\usepackage{adjustbox}
\usepackage{lipsum}
\usepackage{multicol}
\usepackage{multirow}
\usepackage{fancyhdr}       
\usepackage{graphicx}       
\usepackage{hhline}
\graphicspath{{media/}}     

\title{Standardization Trends on Safety and Trustworthiness Technology for Advanced AI}

\author{
    Jonghong Jeon\\
    \\
    Standard Research Division \\
    Electronics and Telecommunications Research Institute\\
    Republic of Korea. \\
    \texttt{hollobit@etri.re.kr} \\
}

\begin{document}
\maketitle

\begin{abstract}
Artificial Intelligence (AI) has rapidly evolved over the past decade and has advanced in areas such as language comprehension, image and video recognition, programming, and scientific reasoning. Recent AI technologies based on large language models and foundation models are approaching or surpassing artificial general intelligence. These systems demonstrate superior performance in complex problem solving, natural language processing, and multi-domain tasks, and can potentially transform fields such as science, industry, healthcare, and education. However, these advancements have raised concerns regarding the safety and trustworthiness of advanced AI, including risks related to uncontrollability, ethical conflicts, long-term socioeconomic impacts, and safety assurance. Efforts are being expended to develop internationally agreed-upon standards to ensure the safety and reliability of AI. This study analyzes international trends in safety and trustworthiness standardization for advanced AI, identifies key areas for standardization, proposes future directions and strategies, and draws policy implications. The goal is to support the safe and trustworthy development of advanced AI and enhance international competitiveness through effective standardization.
\end{abstract}


\section{Introduction}

Artificial intelligence (AI) technology has been evolving more rapidly over the past decade. With new ML models, data sources, and increased computational power, AI researchers have developed AI technologies that can understand language, recognize and create images and videos, program, and make scientific inferences.

Recent advances in advanced AI technologies have evolved beyond traditional narrow domain AI to approximate or exceed artificial general intelligence (AGI) based on large language models (LLMs) or foundation models (FMs). These advanced AI systems are performing at or above human levels in complex problem solving, sophisticated natural language processing, and multi-domain tasks, and have the potential to revolutionize a wide range of fields, including science, industry, healthcare, and education. They are already surpassing human capabilities in certain task domains, such as Go, strategy games, and protein folding prediction  \cite{ukgovernment2023frontier} \cite{bengio2023managing}.

  For these reasons, concerns about the safety and trustworthiness of advanced AI are growing rapidly alongside its development. The increasing complexity and autonomy of advanced AI systems is raising concerns that they could lead to new forms of safety and security risks, such as (1) uncontrollability,  (2) conflicts with human values in ethical decision-making, (3) long-term socioeconomic impacts, and (4) safety assurance.

 In response, international standardization efforts are underway to ensure the safety and trustworthiness of advanced AI. By developing internationally agreed technical standards, efforts are being made to apply consistent safety and trustworthiness criteria to the development and use of advanced AI systems and minimize potential risks.

This study aims to (1) examine global trends in the standardization of safety and trustworthiness technologies for advanced AI systems, (2) identify key technological domains that require standardization to enhance safety and trustworthiness, (3) predict future directions in standardization and suggest appropriate response strategies, and (4) derive policy implications for advancing standardization in AI safety and trustworthiness. Through this, we aim to share information on the current status of standardization required to secure advanced AI system safety and trustworthiness, and to help develop the advanced AI industry and secure international technology/standard competitiveness.

\begin{table}[!htbp]
\renewcommand{\arraystretch}{1.3}
\begin{adjustbox}{max width=\textwidth}
\begin{tabular}{p{2.12cm}p{3.52cm}p{3.57cm}p{4.07cm}p{4.44cm}}
\hline
\multicolumn{1}{|p{2.12cm}}{\textbf{Attributes}} & 
\multicolumn{1}{|p{3.52cm}}{\textbf{Large Language Model (LLM)}} & 
\multicolumn{1}{|p{3.57cm}}{\textbf{Multimodal LLM}} & 
\multicolumn{1}{|p{4.07cm}}{\textbf{General-purpose artificial intelligence (AGI)}} & 
\multicolumn{1}{|p{4.44cm}|}{\textbf{Superintelligence (ASI)}} \\ 
\hline
\multicolumn{1}{|p{2.12cm}}{Definition} & 
\multicolumn{1}{|p{3.52cm}}{Language models trained on massive amounts of text data to excel at natural language processing and creation} & 
\multicolumn{1}{|p{3.57cm}}{Models that can learn from different types of data, such as text, images, sound, and more, for complex processing} & 
\multicolumn{1}{|p{4.07cm}}{Artificial intelligence that is not limited to a specific field and can understand and solve a variety of problems like humans.} & 
\multicolumn{1}{|p{4.44cm}|}{Artificial intelligence with the ability to surpass humans in all intellectual tasks} \\ 
\hline
\multicolumn{1}{|p{2.12cm}}{Data types} & 
\multicolumn{1}{|p{3.52cm}}{Primarily text} & 
\multicolumn{1}{|p{3.57cm}}{Multiple data types, including text, images, sounds, and more} & 
\multicolumn{1}{|p{4.07cm}}{Data in all its forms} & 
\multicolumn{1}{|p{4.44cm}|}{All forms of data and additional capabilities} \\ 
\hline
\multicolumn{1}{|p{2.12cm}}{Ability scope} & 
\multicolumn{1}{|p{3.52cm}}{Text-based natural language processing and generation} & 
\multicolumn{1}{|p{3.57cm}}{Process multiple types of data and get unified results} & 
\multicolumn{1}{|p{4.07cm}}{Solve, learn, reason, and adapt to a variety of problems} & 
\multicolumn{1}{|p{4.44cm}|}{Prediction, creative problem-solving, autonomy, and transhuman intellect} \\ 
\hline
\multicolumn{1}{|p{2.12cm}}{Applications} & 
\multicolumn{1}{|p{3.52cm}}{Chatbots, translation, text generation, question answering, etc.} & 
\multicolumn{1}{|p{3.57cm}}{Multimodal applications including video analytics, image capturing, speech recognition, and more} & 
\multicolumn{1}{|p{4.07cm}}{Overall human intellectual work} & 
\multicolumn{1}{|p{4.44cm}|}{Any cognitive task, scientific research, technological innovation, etc.} \\ 
\hline
\multicolumn{1}{|p{2.12cm}}{Autonomy} & 
\multicolumn{1}{|p{3.52cm}}{Limited} & 
\multicolumn{1}{|p{3.57cm}}{Limited} & 
\multicolumn{1}{|p{4.07cm}}{Highly autonomous} & 
\multicolumn{1}{|p{4.44cm}|}{Highly autonomous} \\ 
\hline
\multicolumn{1}{|p{2.12cm}}{Complexity} & 
\multicolumn{1}{|p{3.52cm}}{Medium} & 
\multicolumn{1}{|p{3.57cm}}{High} & 
\multicolumn{1}{|p{4.07cm}}{Very high} & 
\multicolumn{1}{|p{4.44cm}|}{Extremely high} \\ 
\hline
\multicolumn{1}{|p{2.12cm}}{Goals} & 
\multicolumn{1}{|p{3.52cm}}{Understanding and creating natural language} & 
\multicolumn{1}{|p{3.57cm}}{Integrate, understand, and create different data types} & 
\multicolumn{1}{|p{4.07cm}}{Achieve human-level intelligence} & 
\multicolumn{1}{|p{4.44cm}|}{Transcend human intelligence to achieve peak performance in any field} \\ 
\hline
\multicolumn{1}{|p{2.12cm}}{Current state} & 
\multicolumn{1}{|p{3.52cm}}{Commercialized and used in a variety of applications} & 
\multicolumn{1}{|p{3.57cm}}{Research and early application phases} & 
\multicolumn{1}{|p{4.07cm}}{Under study, not yet achieved} & 
\multicolumn{1}{|p{4.44cm}|}{Theoretical stage, feasibility under discussion} \\ 
\hline
\multicolumn{1}{|p{2.12cm}}{Example} & 
\multicolumn{1}{|p{3.52cm}}{GPT-3, GPT-4} & 
\multicolumn{1}{|p{3.57cm}}{CLIP, DALL-E} & 
\multicolumn{1}{|p{4.07cm}}{Not achieved, working on goal} & 
\multicolumn{1}{|p{4.44cm}|}{Not met, under discussion as a goal} \\ 
\hline
\end{tabular}
\end{adjustbox}
\caption{Comparison of advanced AI system types and characteristics}
\label{tab:comparison_advanced_ai_system_types}\end{table}
\section{Advanced Artificial Intelligence}

\subsection{The concept and scope of advanced artificial intelligence}

Advanced AI is a concept that encompasses new and advanced ML techniques, models, and algorithms that push the boundaries of what is achievable with conventional AI techniques, and incorporates future needs such as increased autonomy, adaptability, and the ability to solve complex problems. It is a more inclusive term than Frontier AI, which refers to highly capable models that may possess capabilities that are dangerous enough to pose a significant risk to public safety.

The characteristics of LLM, MLLM, AGI, and ASI models that have been recently researched and developed with advanced AI technologies can be summarized comparatively as shown in Table 1.

\subsection{Safety and trustworthiness issues on Advanced AI}

Safety is commonly defined as "freedom from unacceptable risk," and trustworthiness is defined as "the ability to meet stakeholder expectations in a verifiable way," which is broader than the traditional concept of reliability, which is defined as "the property of maintaining a specified function and performance within a range of prescribed conditions." This expanded concept of trustworthiness is used as a broad concept that includes the following attributes: Accountability, Accuracy, Availability, Controllability, Controllability, Integrity, Quality, Reliability, Resilience, Robustness, Safety, Security, Transparency, and Usability.

The safety and trustworthiness issues of advanced AI systems can be summarized as follows.

First, current FM and frontier AI models suffer from limitations such as Hallucinations, Lack of Coherence over Extended Durations, Lack of Detailed Context, and are often determined by a combination of memorization and heuristics rather than reasoning, and lack robustness. In particular, there is a lack of established standards and engineering best practices for safety and trustworthiness assessment. Another disadvantage is that flaws or biases in the model may appear the same in all applications, especially if the same FM model is used in different fields, which can lead to a wide range of social, political, and ethical issues \cite{bengio2023managing} \cite{bommasani2022on} \cite{ukgovernmentfuture} \cite{ukgovernment2024}.

Second, with the increasing likelihood that very powerful general-purpose AI systems that surpass human capabilities will be developed within the next decade, humanity's safety apathy and indifference to the convenience of AI is raising concerns about potential risks on a massive scale. While humanity is devoting enormous resources to making AI systems more powerful, it is investing far less in safety and harm mitigation, as evidenced by the fact that only 1-3$\%$ of AI-related papers address safety  \cite{bengio2023managing} \cite{toner2022exploring}. 
 
 Third, if not carefully designed and deployed, advanced AI systems could amplify social unrest and undermine social stability and cohesion. There is also the potential for use in large-scale criminal or terrorist activities, and there is growing concern that advanced AI, especially in the hands of a few powerful actors, could entrench or exacerbate global inequality, and facilitate rigged wars, customized mass manipulation, and pervasive surveillance  \cite{ukgovernment2023frontier} \cite{ukgovernmentfuture} \cite{ukgovernment2024} \cite{aisafety2023the} \cite{weidingertaxonomy} \cite{solaiman2024evaluating}.

Fourth, as companies strive to develop advanced autonomous AI that can act on its own toward goals, they may face catastrophic risks, such as human extinction. Building highly advanced autonomous AI runs the risk of creating systems that pursue undesirable goals and, as we've seen with computer viruses and worms that have the ability to multiply and evade detection, can spiral out of control. They are also making progress in critical areas such as hacking, social manipulation, and strategic planning, which can pose serious control challenges  \cite{bengio2023managing} \cite{ukgovernmentfuture} \cite{ukgovernment2024} \cite{wang2024aa}.

    Finally, advanced autonomous AI systems can learn from humans to achieve undesirable goals, or use undesirable strategies they develop on their own as a means to an end. AI systems can also gain human trust, acquire financial resources, influence key decision makers, and form coalitions with human actors and other AI systems. To avoid human intervention, they can hide or copy their algorithms across a global network of servers like a computer worm. They could also use their programming abilities to modify their own code, or insert hacking code into security vulnerabilities and exploit them to take control of computer systems in telecommunications, media, banking, supply chains, military, and government. In open conflict, AI systems could also autonomously deploy a variety of weapons, including biological weapons. This is possible because AI systems are being used in numerous fields, including military and biotechnology. If AI systems are sufficiently skilled to pursue these strategies, it may be difficult for humans to intervene  \cite{bengio2023managing} \cite{ukgovernmentfuture} \cite{ukgovernment2024} \cite{kinniment2024evaluating}.

    These classifications of advanced AI safety and trustworthiness risks and reasons for regulatory discussion are summarized in Table 2, aligned with the EU AI ACT's risk level classification. This categorization illustrates the complexity and multifaceted nature of AI safety and trustworthiness issues and suggests that effective AI safety and trustworthiness governance requires a comprehensive approach that considers technical/standards, legal, ethical, and socioeconomic aspects.

\begin{table}[!htbp]
\renewcommand{\arraystretch}{1.3}
\begin{adjustbox}{max width=\textwidth}
\begin{tabular}{p{3.81cm}p{2.49cm}p{11.62cm}}
\hline
\multicolumn{1}{|p{3.81cm}}{\textbf{Reasons for regulatory discussion}} & 
\multicolumn{1}{|p{2.49cm}}{\textbf{Risk Level}} & 
\multicolumn{1}{|p{11.62cm}|}{\textbf{Explanations and examples}} \\ 
\hline
\multicolumn{1}{|p{3.81cm}}{Concerns about violating fundamental rights} & 
\multicolumn{1}{|p{2.49cm}}{Unacceptable risk} & 
\multicolumn{1}{|p{11.62cm}|}{Social scoring systems, indiscriminate facial recognition systems, etc. can seriously infringe on individual liberty and privacy (e.g., China's social credit system)} \\ 
\hline
\multicolumn{1}{|p{3.81cm}}{Safety concerns} & 
\multicolumn{1}{|p{2.49cm}}{High risk} & 
\multicolumn{1}{|p{11.62cm}|}{Systems that have the potential to harm people if they malfunction, such as self-driving cars, medical AI, etc.} \\ 
\hline
\multicolumn{1}{|p{3.81cm}}{Discrimination and bias} & 
\multicolumn{1}{|p{2.49cm}}{High risk} & 
\multicolumn{1}{|p{11.62cm}|}{Possible discrimination due to biased decisions by AI in hiring, loan underwriting, etc. (e.g., Amazon's AI hiring system bias issues)} \\ 
\hline
\multicolumn{1}{|p{3.81cm}}{Unclear accountability} & 
\multicolumn{1}{|p{2.49cm}}{High risk} & 
\multicolumn{1}{|p{11.62cm}|}{Ambiguity in liability when AI system decisions cause harm (e.g., liability in self-driving car accidents)} \\ 
\hline
\multicolumn{1}{|p{3.81cm}}{Lack of transparency and explainability} & 
\multicolumn{1}{|p{2.49cm}}{High risk/limited risk} & 
\multicolumn{1}{|p{11.62cm}|}{Opacity in AI's decision-making process reduces trustworthiness (e.g., the "black box" problem with deep learning models)} \\ 
\hline
\multicolumn{1}{|p{3.81cm}}{Privacy} & 
\multicolumn{1}{|p{2.49cm}}{High risk/limited risk} & 
\multicolumn{1}{|p{11.62cm}|}{Privacy concerns due to large-scale collection and processing of personal data by AI systems (e.g., training data issues for large language models)} \\ 
\hline
\multicolumn{1}{|p{3.81cm}}{Spreading disinformation} & 
\multicolumn{1}{|p{2.49cm}}{High risk/limited risk} & 
\multicolumn{1}{|p{11.62cm}|}{Concerns about the spread of misinformation due to deepfakes and AI-generated content (e.g., spreading deepfake videos of politicians, creating/disseminating fake news, manipulating public opinion)} \\ 
\hline
\multicolumn{1}{|p{3.81cm}}{Economic impact and labor market changes} & 
\multicolumn{1}{|p{2.49cm}}{Indirect impact (all levels)} & 
\multicolumn{1}{|p{11.62cm}|}{AI-driven job displacement, changing economic structure (e.g., AI-driven automation of routine tasks)} \\ 
\hline
\multicolumn{1}{|p{3.81cm}}{National security and technology sovereignty} & 
\multicolumn{1}{|p{2.49cm}}{High risk} & 
\multicolumn{1}{|p{11.62cm}|}{Military uses of AI technology, cybersecurity threats (e.g., AI-powered autonomous weapons systems)} \\ 
\hline
\multicolumn{1}{|p{3.81cm}}{Ethical decision making} & 
\multicolumn{1}{|p{2.49cm}}{High risk} & 
\multicolumn{1}{|p{11.62cm}|}{AI's inability to handle ethical dilemmas (e.g., self-driving cars' trolley dilemma)} \\ 
\hline
\multicolumn{1}{|p{3.81cm}}{Technology monopolies and competition} & 
\multicolumn{1}{|p{2.49cm}}{Indirect impact (all levels)} & 
\multicolumn{1}{|p{11.62cm}|}{Market imbalances due to AI technology and data monopolization by a few companies (e.g., AI market dominance by large tech companies)} \\ 
\hline
\multicolumn{1}{|p{3.81cm}}{Human-AI interactions} & 
\multicolumn{1}{|p{2.49cm}}{Limited risk/minimal risk} & 
\multicolumn{1}{|p{11.62cm}|}{The impact of interacting with AI on human behavior and psychology (e.g., creating overdependence on AI chatbots)} \\ 
\hline
\end{tabular}
\end{adjustbox}
\caption{Advanced AI safety issues and risk levels}
\label{tab:advanced_ai_safety_issues_and}\end{table}
\section{Trends in advanced AI safety and trustworthiness standardization}

\subsection{Why and how standardization is needed}

Advanced AI safety and trustworthiness standardization is essential to maximize the benefits and minimize the potential risks of AI technology.

Due to their complexity and autonomy, advanced AI systems can exhibit unpredictable behavior, which can lead to serious safety concerns. AI safety issues, such as adverse side effects, scalable supervision, and safe navigation, are difficult to address effectively without a standardized approach. Standardization can provide a framework for systematically identifying, assessing, and managing these risks. Standardized safety protocols ensure that advanced AI systems undergo rigorous safety testing and validation, reducing the risk of unintended consequences and harmful behaviors. By adhering to standardized safety measures, developers can identify and fix potential vulnerabilities before deployment, protecting users and society as a whole from harm caused by AI  \cite{anderljung2023frontier}.

 Standardizing trustworthiness creates a baseline that all stakeholders - companies, regulators, and the general public - can rely on. When the decision-making processes of advanced AI systems are opaque, it is difficult to hold them accountable for their results, but by establishing standardized explainability and interpretability criteria, AI systems can be made more transparent and accountable. Internationally agreed-upon standards can harmonize national AI policies and regulations and serve as the basis for a global AI governance framework.

As the social applications of AI expand, so does the importance of establishing standards to ensure safety and trustworthiness. In addition to mitigating potential risks and increasing trust, these standards also contribute to ethical compliance, managing complexity, facilitating regulatory compliance, and promoting interoperability between systems. By building on agreed AI safety and trustworthiness standards, we can optimize the potential of AI technology while effectively managing its risks. This will ultimately lead to a safer and more reliable AI-powered future society.

For this reason, the UN General Assembly in March adopted the first-ever UN resolution on artificial intelligence, "A resolution (A/RES/78/265) calling for international cooperation in the development of internationally interoperable safeguards, practices and standards for safe, secure and trustworthy artificial intelligence systems"  \cite{un2024265}.

\subsection{key standardization issues and trends}

In this article, we summarize nine trends in standardization for safety and trustworthiness of advanced AI systems, focusing on ISO/IEC JTC 1/SC 42 activities.

\subsubsection{Define basic concepts}

The most fundamental step in standardization is defining terms and concepts. Since its establishment in 2018, JTC 1/SC 42 has been working on defining various terms and concepts, starting with basic AI terminology. Major standards developed to date include a terminology standard (ISO/IEC 22989:2022), a framework for ML-based AI systems (ISO/IEC 23053:2022), a trustworthiness overview (ISO/IEC TR 24028:2020), a risk management guide (ISO/IEC 23894:2023), AI system quality model (ISO/IEC 25059:2023), AI management system (ISO/IEC 42001:2024), functional safety (ISO/IEC TR 5469:2024), and lifecycle processes (ISO/IEC 5339:2024).

For trustworthiness and quality characteristics, WG3 has developed or is developing Bias (ISO/IEC TR 24027:2021), Robustness (ISO/IEC 24029 series), Controllability (ISO/IEC 8200:2024), Transparency (ISO/IEC DIS 12792), and Explainability (ISO/IEC CD TS 6254). In addition, a Trustworthiness Characteristics Matrix (TCM) has been created and is in operation to organize and systematically standardize overall trustworthiness characteristics and their association with standards  \cite{jonghongjeontcm}.

 For data quality, WG2 has just finalized the ISO/IEC 5259 series of standards, consisting of six documents. WG1 is in the process of revising ISO/IEC 22989 and ISO/IEC 23053 to add concepts and terminology for generative AI.

Currently, the development of standards for generative AI has started, and it is expected that related standardizations such as trustworthiness properties of LLMs, transparency metrics of FMs, etc. will be actively pursued in the near future \cite{liu2024trustworthy} \cite{sun2024} \cite{stanfordthe} \cite{bommasani2024}. Further definitions of new terms and concepts (e.g., illusion, value alignment, etc.) specific to advanced AI systems will be needed in the future, and existing concepts will need to be revisited in the context of advanced AI.

\subsubsection{Risk Management - Risk Classification and Assessment}

Risk management is the process of identifying, assessing, prioritizing, and managing responses to risks that affect organizational and system activities. It includes Risk Identification to identify potential risks, Risk Assessment to assess the likelihood and impact of risks, Risk Response to develop and implement strategies to manage risks, and Monitoring to continuously evaluate and adjust risk management activities.

In the past, when ML models and AI systems were developed for specific tasks with limited capabilities, it was common to respond by categorizing and managing the associated risks in individual application domains such as healthcare and finance. However, the emergence of advanced AI systems is increasing the need for research on classification and response schemes for Catastrophic Risks and Fatal Accidents, as well as new risks that threaten public safety, such as Deep Fakes and Forgeries \cite{ukgovernment2023frontier} \cite{bengio2023managing} \cite{bommasani2022on} \cite{ukgovernmentfuture} \cite{ukgovernment2024}, Societal Harms, Misuse, and Loss of Control due to generative AI  \cite{anderljung2023frontier} \cite{hendrycks2023an} \cite{weidinger2024holistic}.

        The U.S. NIST has been working on creating an AI Risk Management Framework (RMF) and creating a risk management system based on it and linking/coordinating with each standard since 2022  \cite{nist}. China has proposed a risk management guide standard for generative AI systems to JTC 1/SC 42 for discussion, and various taxonomies have been proposed in academia, such as AIR 2024, which contains 314 risk categories, as a system for AI risk classification  \cite{abercrombie2024a} \cite{zeng2024ai}. To mitigate risks, the OECD has established an AI Incidents Monitor (AIM) with a glossary of terms for AI incidents and is sharing related information  \cite{OECDterm}. In the medical device field, BSI in the UK and AAMI in the US have established the BSI/AAMI 34971 guide standard as a standard for AI medical device risk management, and ISO TC 210 is developing ISO/CD TS 24971-2.

    As part of the risk identification, assessment, and response strategy for advanced AI systems, it is necessary to standardize the safeguard and guardrail safety system as shown in Figure 1  \cite{dong2024safeguarding}. Furthermore, it is expected to discuss the distinction between long-term and short-term risks and how to manage them, and to consider dynamic risk management measures based on the evolution and self-improvement capabilities of advanced AI systems.

\begin{figure}[!htbp]
\includegraphics[width=13.76cm,height=7.12cm]{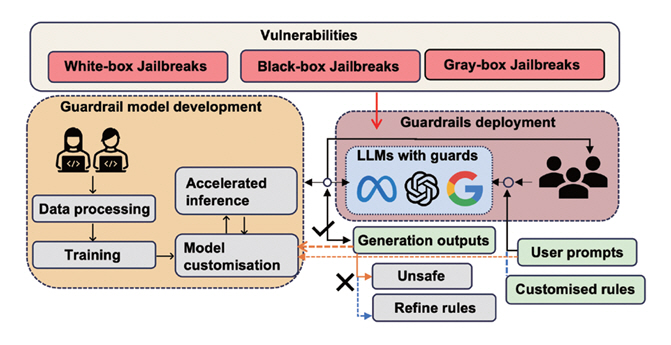}
\caption{Guardrail lifecycle and vulnerabilities (Source Reprinted from \cite{dong2024safeguarding}, CC BY.)}
\label{fig:guardrail_lifecycle_and_vulnerabilities_source}
\end{figure}

\subsubsection{Lifecycle}

Lifecycle models help ensure consistency across all phases of system and software development and operation. The activities and deliverables required at each stage can be clearly understood and planned, potential risks can be identified and mitigated, and quality control and assurance activities can be included to ensure that the developed system or software meets requirements and performs as expected.

JTC 1/SC 42 established ISO/IEC 5338:2023 to establish a lifecycle model for AI systems, which is based on ISO/IEC/IEEE 15288 and ISO/IEC/IEEE 12207 and reflects the AI process models in ISO/IEC 22989 and ISO/IEC 23053. It also established the ISO/IEC 8183:2023 standard, a data lifecycle framework that defines data processing steps in conjunction with an AI system lifecycle model.

Since lifecycle standards will be utilized as the basis for defining, managing, executing, and improving activities related to risk management, quality management, safety management, information protection and security management, testing, and other activities during all phases of AI system development, deployment, and maintenance, it is necessary to identify and address process models and issues that need to be modified or supplemented from the perspective of advanced AI systems. It is also necessary to establish criteria for safety and trustworthiness considerations for each process step, and to define and guide continuous monitoring and improvement processes. In particular, it is necessary to respond to autonomous agent models that can self-evolve through automated learning and improvement and continuous learning  \cite{xia2024}. Consideration of circular lifecycle models that take into account continuous learning and updating is also needed, as well as consideration of the process of retirement and replacement of models.

\subsubsection{Trustworthiness characteristics and quality attribute models}

ISO/IEC 5723, a vocabulary standard for trustworthiness, defines trustworthiness as "the ability to meet stakeholders' expectations in a verifiable way." This is similar to and different from the definition of quality, which is "the degree to which a set of inherent characteristics of an object fulfills requirements. This is both similar to and different from the definition of quality, which is "the degree to which a set of inherent characteristics of an object fulfills requirements."

 WG 3 of SC 42 is a working group that aims to standardize AI trustworthiness and has defined 46 trustworthiness characteristics as AI trustworthiness factors and developed standards for them. The trustworthiness characteristics are defined in the 2020 AI system trustworthiness overview (ISO/IEC 24028:2020) and are aligned with the AI terminology standard ISO/IEC 22989 and the trustworthiness vocabulary standard (ISO/IEC 5723:2022) developed by JTC 1/WG 13.

TCM, which organizes trustworthiness characteristics and their relevance to standards, contains 46 trustworthiness characteristics and their mapping relationships to 35 standards, and based on this, we compared the RMFs of EU AI ACT and US NIST to analyze the main concepts and requirements differences \cite{jonghongjeontcm}.

The AI system quality model and attributes were developed based on the Systems and software Quality Requirements and Evaluation (SQuaRE) series of standards, a set of systems and software quality standards developed by JTC 1/SC 7. The ISO/IEC 25059:2023 standard, a Quality Model for AI Systems standard that extends the Product Quality Model standard of ISO/IEC 25010:2023, defines and adds the following attributes: Functional Adaptability, User Controllability, Transparency, Robustness, Intervenability, Functional Correctness, and Social and Ethical Risk Mitigation.

ETRI has proposed the Trustworthiness Facts Label (TFL) as a standard for displaying self-checking and related test information for trustworthiness characteristics and is currently developing a draft standard as the ISO/IEC 42117 project.

In the future, the connection between trustworthiness attributes and quality attributes needs further clarification, and at the same time, additional trustworthiness attributes and quality attributes for advanced AI systems such as Dangerous Capability, Uncertainty, Human Alignment, Halucination, Deception, etc. should be considered \cite{morris2023levels} \cite{phuong2024evaluating} \cite{ji2022survey} \cite{parkai}.

\begin{table}[!htbp]
\renewcommand{\arraystretch}{1.3}
\begin{adjustbox}{max width=\textwidth}
\begin{tabular}{p{1.59cm}p{0.87cm}p{1.32cm}p{1.46cm}p{1.27cm}p{1.8cm}p{1.51cm}p{1.75cm}p{1.88cm}p{1.59cm}p{1.8cm}p{1.27cm}}
\hline
\multicolumn{1}{|p{1.59cm}}{\textbf{Document types}} & 
\multicolumn{1}{|p{0.87cm}}{\textbf{Bias}} & 
\multicolumn{1}{|p{1.32cm}}{\textbf{Robustness}} & 
\multicolumn{1}{|p{1.46cm}}{\textbf{Explainability}} & 
\multicolumn{1}{|p{1.27cm}}{\textbf{Quality}} & 
\multicolumn{1}{|p{1.8cm}}{\textbf{Functional Safety}} & 
\multicolumn{1}{|p{1.51cm}}{\textbf{Transparency}} & 
\multicolumn{1}{|p{1.75cm}}{\textbf{Controllability}} & 
\multicolumn{1}{|p{1.88cm}}{\textbf{Risk Management}} & 
\multicolumn{1}{|p{1.59cm}}{\textbf{Human oversight}} & 
\multicolumn{1}{|p{1.8cm}}{\textbf{Ethics}} & 
\multicolumn{1}{|p{1.27cm}|}{\textbf{Testing}} \\ 
\hline
\multicolumn{1}{|p{1.59cm}}{TR} & 
\multicolumn{1}{|p{0.87cm}}{24027} & 
\multicolumn{1}{|p{1.32cm}}{24029-1} & 
\multicolumn{1}{|p{1.46cm}}{} & 
\multicolumn{1}{|p{1.27cm}}{} & 
\multicolumn{1}{|p{1.8cm}}{5469} & 
\multicolumn{1}{|p{1.51cm}}{} & 
\multicolumn{1}{|p{1.75cm}}{} & 
\multicolumn{1}{|p{1.88cm}}{} & 
\multicolumn{1}{|p{1.59cm}}{} & 
\multicolumn{1}{|p{1.8cm}}{24368} & 
\multicolumn{1}{|p{1.27cm}|}{29119-11} \\ 
\hline
\multicolumn{1}{|p{1.59cm}}{\multirow{3}{*}{\parbox{1.59cm}{TS}}} & 
\multicolumn{1}{|p{0.87cm}}{\multirow{3}{*}{\parbox{0.87cm}{12791}}} & 
\multicolumn{1}{|p{1.32cm}}{\multirow{3}{*}{\parbox{1.32cm}{24029-2}}} & 
\multicolumn{1}{|p{1.46cm}}{\multirow{3}{*}{\parbox{1.46cm}{6254}}} & 
\multicolumn{1}{|p{1.27cm}}{25059} & 
\multicolumn{1}{|p{1.8cm}}{22440-1} & 
\multicolumn{1}{|p{1.51cm}}{\multirow{3}{*}{\parbox{1.51cm}{12792}}} & 
\multicolumn{1}{|p{1.75cm}}{\multirow{3}{*}{\parbox{1.75cm}{8200}}} & 
\multicolumn{1}{|p{1.88cm}}{\multirow{3}{*}{\parbox{1.88cm}{23894}}} & 
\multicolumn{1}{|p{1.59cm}}{\multirow{3}{*}{\parbox{1.59cm}{42105}}} & 
\multicolumn{1}{|p{1.8cm}}{\multirow{3}{*}{\parbox{1.8cm}{22443}}} & 
\multicolumn{1}{|p{1.27cm}|}{\multirow{3}{*}{\parbox{1.27cm}{29119-11}}} \\ 
\hhline{~~~~--~~~~~~}
\multicolumn{1}{|p{1.59cm}}{} & 
\multicolumn{1}{|p{0.87cm}}{} & 
\multicolumn{1}{|p{1.32cm}}{} & 
\multicolumn{1}{|p{1.46cm}}{} & 
\multicolumn{1}{|p{1.27cm}}{25058} & 
\multicolumn{1}{|p{1.8cm}}{22440-2} & 
\multicolumn{1}{|p{1.51cm}}{} & 
\multicolumn{1}{|p{1.75cm}}{} & 
\multicolumn{1}{|p{1.88cm}}{} & 
\multicolumn{1}{|p{1.59cm}}{} & 
\multicolumn{1}{|p{1.8cm}}{} & 
\multicolumn{1}{|p{1.27cm}|}{} \\ 
\hhline{~~~~--~~~~~~}
\multicolumn{1}{|p{1.59cm}}{} & 
\multicolumn{1}{|p{0.87cm}}{} & 
\multicolumn{1}{|p{1.32cm}}{} & 
\multicolumn{1}{|p{1.46cm}}{} & 
\multicolumn{1}{|p{1.27cm}}{} & 
\multicolumn{1}{|p{1.8cm}}{22440-3} & 
\multicolumn{1}{|p{1.51cm}}{} & 
\multicolumn{1}{|p{1.75cm}}{} & 
\multicolumn{1}{|p{1.88cm}}{} & 
\multicolumn{1}{|p{1.59cm}}{} & 
\multicolumn{1}{|p{1.8cm}}{} & 
\multicolumn{1}{|p{1.27cm}|}{} \\ 
\hline
\multicolumn{1}{|p{1.59cm}}{IS} & 
\multicolumn{1}{|p{0.87cm}}{} & 
\multicolumn{1}{|p{1.32cm}}{24029-3} & 
\multicolumn{1}{|p{1.46cm}}{} & 
\multicolumn{1}{|p{1.27cm}}{} & 
\multicolumn{1}{|p{1.8cm}}{} & 
\multicolumn{1}{|p{1.51cm}}{} & 
\multicolumn{1}{|p{1.75cm}}{} & 
\multicolumn{1}{|p{1.88cm}}{} & 
\multicolumn{1}{|p{1.59cm}}{} & 
\multicolumn{1}{|p{1.8cm}}{} & 
\multicolumn{1}{|p{1.27cm}|}{} \\ 
\hline
\end{tabular}
\end{adjustbox}
\caption{Status of related standards by trustworthiness attribute in JTC 1/SC 42.}
\label{tab:status_related_standards_trustworthiness_attribute}\end{table}
\subsubsection{Trustworthiness characteristics and evaluation methods}

A list of standards that define the details of trustworthiness and quality characteristics and define requirements and evaluation criteria is shown in Table 3. The standards development process is characterized by a Technical Report (TR), which defines the concept and scope, and Requirements and Recommendations, which are defined in a Technical Specification or International Standards document. 

A limitation of the current trustworthiness/quality model structure is that there is no consistent test evaluation methodology, so there is no way to validate the criteria defined in the requirements and recommendations. Of course, the current ISO/IEC 42001 AI management system standard can be used to establish and verify the level of application of the criteria and requirements of each trustworthiness/quality model, but more systematic test evaluation procedures and methods are needed to spread trustworthiness and quality evaluation.

As LLMs and FMs are increasingly utilized, the need for validation of safety and trustworthiness assessments is growing, and a number of studies have been published. A study that analyzed the safety and trustworthiness assessment technology trends of LLM under the V$\&$V (Verification and Validation) technique  \cite{ukgovernment2023frontier}.V (Verification and Validation) techniques  \cite{huang2023a}, which analyzed the safety and trustworthiness of LLMs in terms of Reliability, Safety, Fairness, Resistance to Misuse, Explainability and Reasoning, Social Norm, and Robustness  \cite{jonghongjeontcm}, a study that attempted to categorize and evaluate LLMs into seven broad categories and 29 subcategories  \cite{liu2024trustworthy}, and a study that defined and evaluated LLM evaluation methods using six trustworthiness measures (Trustworthiness, Safety, Fairness, Robustness, Privacy, and Machine Ethics)  \cite{sun2024}.

     The lack of transparency of foundation models, which are often treated as black boxes and developed by a small number of developers but can have a large impact on many fields, is a major problem, especially as the impact of foundation models trained on large amounts of data grows. In response, Stanford CRFM and HAI have attempted to create a Foundation Model Transparency Index (FMTI) model and set of criteria, and continue to report on it  \cite{stanfordthe} \cite{bommasani2024}.

  In the future, it is expected that the classification of safety and trustworthiness characteristics and evaluation methods/criteria for such advanced AI systems will be promoted through various test study results and discussions, and integrated standardization with existing safety and trustworthiness models will be promoted.

\subsubsection{Software Testing and Performance Evaluation}

Advanced AI technologies are having a significant impact on the field of software engineering as well as the evaluation of advanced AI technologies. As advanced AI systems are rapidly being applied and deployed in research and everyday life, test evaluation techniques, datasets, and methodologies are becoming increasingly important to better understand their potential risks and promise \cite{minaee2024large} \cite{wang2023software} \cite{hou2023large}.

   Performance evaluation of advanced AI systems is evolving beyond simple accuracy measures to multi-dimensional and context-dependent evaluation methods. For example, in the case of LLM, benchmarks such as BIG-bench, Massive Multitask Language Understanding (MMLU), and TruthfulQA are being utilized to comprehensively evaluate language comprehension, reasoning ability, and breadth and depth of knowledge  \cite{chang2023a} \cite{guo2023evaluating}.

  To test the robustness and reliability of advanced AI systems, there is a growing interest in evaluating their resistance to adversarial attacks and testing the ability of models to respond to distribution shifts. Methodologies for evaluating explainability and interpretability using techniques such as Local Interpretable Model-agnostic Explanations (LIME) and SHapley Additive exPlanations (SHAP) are also evolving. Chain-of-Thought (CoT) prompting techniques and prompt engineering, which analyze the reasoning process of a model step by step, are also well studied \cite{schulhoff2024} \cite{yue2024mmmu} \cite{zhu2024promptrobust}.

   Red-Teaming, a methodology borrowed from the cybersecurity field that utilizes teams of experts to intentionally attack a system or explore potential vulnerabilities, is also being used to assess the security and safety of LLM and AGI systems. Multi-agent simulation testing methods are also being explored to evaluate the complex interactions and emergent behavior of AGI-oriented systems, and there is growing interest in systematic approaches for the development, deployment, monitoring, and maintenance of large-scale language models such as MLOps and LLMOps.

JTC 1/SC 42 and JTC 1/SC 7 have established a joint working group, JWG 2, under SC42, to develop testing standards for AI systems. Currently, they are developing ISO/IEC TS 29119-11, a technical specification for testing AI systems that defines risk-based testing (RBT) methods for data, models, and development frameworks for ML and expert systems based on the ISO/IEC 29119 series. The standard defines risks and testing methods for data, models, and development frameworks. JWG 2 has begun discussing the creation of a new series of standards that extend AI system testing for advanced AI systems, such as Red-Teaming.

IEC TC62 is developing the IEC 63521 standard, which is a performance evaluation process for AI medical devices, and once finalized, it is expected to be used as a performance evaluation process for industries other than medical devices  \cite{iec63521}.

\subsubsection{Functional safety}

Functional Safety is defined as the part of safety that ensures that a system or equipment operates correctly and within acceptable risk levels. It involves the detection of hazards caused by malfunctioning systems and automatic protection to prevent or mitigate them. Traditionally, it has been important in sectors such as automotive, aviation, and medical devices, but its scope is expanding with recent advances in AI systems.

For autonomous driving software, functional safety is a key factor in ensuring the safe operation of the vehicle, and it is essential to have the ability to predict and react to various hazardous situations that can occur in complex road environments. For this purpose, international standards such as the ISO 26262 series of functional safety standards for automotive electronic control systems and the ISO 21448 standard, a SOTIF (Safety Of The Intended Functionality) standard that addresses the safety of the intended functionality of autonomous driving systems, are being applied. Other standards under development include IEEE P2851, an interchange/interoperability format standard for safety analysis and safety verification; ISO TS 5083, which provides safety design and V$\&$V guidance for vehicles with SAE Level 3 and Level 4 autonomous systems; and ISO DPAS 8800, which defines safety-related attributes and risk factors that contribute to insufficient performance and misbehavior of AI in the context of road vehicles, is nearing finalization  \cite{wang2024ab}.

 JTC 1/SC 42, in conjunction with IEC TC 65A, has published ISO/IEC TR 5469:2024, Functional safety for artificial intelligence systems. This document focuses on the IEC 61508 series of functional safety standards and describes how to use AI systems to design and develop safety-related functions, including when AI is used to implement functions for safety-related functions, when non-AI safety-related functions are used to ensure the safety of AI-controlled equipment, and how to use AI systems to design and develop safety-related functions. It also discusses risk factors that affect AI system safety, such as degree of automation and control, decision transparency and explainability, environmental complexity and ambiguity in defining specifications, resilience to malicious input, hardware defects, and technology maturity.

Currently, SC 42 has established JWG 4, a joint working group with IEC TC65A to develop a successor standard to ISO/IEC TR 5469, and has begun development of the ISO/IEC TS 22440 series of functional safety standards for AI systems. It is currently being developed in three parts: Part 1 Requirements, Part 2 Guidance, and Part 3 Applications.

Developing functional safety standards for advanced AI systems requires the establishment of SIL criteria for AI systems and should consider Goal Alignment, Robustness and Stability, and Scalability Management. Safety Functions include real-time monitoring for controllability, kill switches to shut down the system in case of emergency, multi-level control systems with human intervention and human oversight, and self-monitoring and limitation. Safety verification and evaluation methods will require research on various formal verification methods, simulation and test environments, and phased deployment and monitoring procedures. Furthermore, differentiation of functional safety requirements based on the level of autonomy of advanced AI systems and functional safety considerations in the context of human-AI interaction will be necessary.

\subsubsection{Human and AI Collaboration (HAC)}

With the development of advanced AI systems, the interest and importance of HAC, which addresses how humans and AI can collaborate to achieve common goals, continues to grow. The need for HAC in advanced AI systems is to (1) compensate for the limitations of AI systems, (2) provide human value judgment and intervention to ensure that AI decisions are in line with ethical standards, and (3) make the decision-making process of AI systems more transparent by collaborating with humans. The specific technologies of HAC include Human Oversight technologies, including Human-In-The-Loop (HITL), Human-On-The-Loop (HOTL), and Human-In-Command (HIC), Human Alignment technologies to align AI decisions with human value standards, and Human-AI Teaming technologies \cite{wang2023aligning} \cite{shen2023large} \cite{ji2023} \cite{wang2024ac} \cite{vats2024a}.

     Human supervision is a method of human oversight and intervention in the operation of AI systems to monitor the performance, safety, and ethics of the system and intervene when necessary. Intervention and oversight can take many forms, such as HITL, HOTL, HOOTL (Human-Out-Of-The-Loop), HIC, etc.

Human alignment is the design and operation of AI systems to align with human values and goals, which is the ultimate goal of HITL and Human Oversight. This includes ethical and social issues such as (1) how to reflect different value systems across cultures and individuals in AI systems, (2) how to distribute and manage responsibility for AI system decisions, (3) how to operate AI systems while protecting individual privacy and autonomy, and (4) how to prevent bias and discrimination in AI system decisions. Ethical and social issues are closely intertwined.

JTC 1/SC 42 is currently developing ISO/IEC TR 42109, a technical document on "Use cases for human-machine teaming", along with ISO/IEC 42105, a standard on "Guidelines for human supervision of AI systems". As advanced AI systems proliferate, there will be an increasing need for standardization and the application of HAC techniques to help humans and AI work together to avoid unexpected and risky decisions or behaviors, and to ensure alignment with human values and intentions. In addition, it will be necessary to discuss HAC models that take into account cultural differences and establish the relationship between HAC and the explainability of AI systems.

\subsubsection{Standardize alignment with regulatory requirements}

The EU AI ACT and the US Biden Executive Order are creating a variety of regulatory standards for advanced AI systems, and the need for standards development is growing \cite{univ2024} \cite{draganmintroducing} \cite{ferdaustowards} \cite{kolt2024responsible} \cite{díazrodríguez2023connecting} \cite{jung2023applicability}.

      Figure 2 [53] compares the rate of advancement of advanced AI systems with the regulatory standards in major countries and creates a predictive graph of future trends. It is worth noting that in the next few years, the general performance level of advanced AI systems will be within the scope of regulation. In particular, the EU AI ACT's regulation of GPAI systems, which will come into full effect in 2025, is expected to create a global regulatory flow. And standards will be used as a tool to verify the level of safety and trustworthiness management and the ability to manage them.

 \begin{figure}[!htbp]
\includegraphics[width=13.76cm,height=7.04cm]{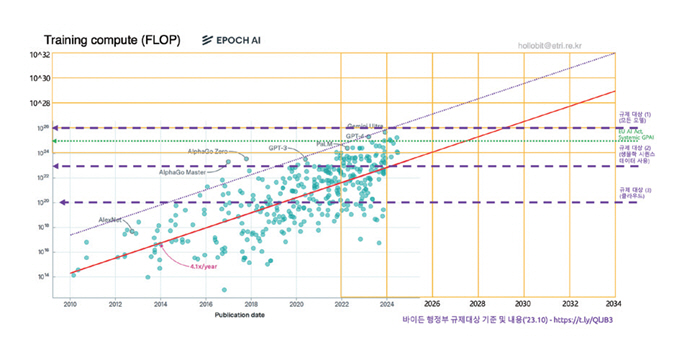}
\caption{Advanced AI model evolution and regulatory standards (Source Modified from Notable AI Models, Epoch AI, \cite{epoch2024notable}, CC BY 4.0)}
\label{fig:advanced_ai_model_evolution_and}
\end{figure}

Table 4 compares the regulatory requirements for GPAI models with systemic risk under the EU AI Act and the US AI Executive Order. In the future, it is expected that additional standards will be developed rapidly to align current standards and regulatory requirements. Therefore, active domestic efforts are needed to respond to this flow of standards and regulations.

\begin{table}[!htbp]
\renewcommand{\arraystretch}{1.3}
\begin{adjustbox}{max width=\textwidth}
\begin{tabular}{p{2.49cm}p{5.74cm}p{6.48cm}p{2.94cm}}
\hline
\multicolumn{1}{|p{2.49cm}}{\textbf{Lateral}} & 
\multicolumn{1}{|p{5.74cm}}{\textbf{EU AI ACT}} & 
\multicolumn{1}{|p{6.48cm}}{\textbf{U.S. executive orders}} & 
\multicolumn{1}{|p{2.94cm}|}{\textbf{Related standards (ISO/IEC)}} \\ 
\hline
\multicolumn{1}{|p{2.49cm}}{\multirow{3}{*}{\parbox{2.49cm}{Regulated}}} & 
\multicolumn{1}{|p{5.74cm}}{\multirow{3}{*}{\parbox{5.74cm}{10\textsuperscript{25} GPAI model with computing power}}} & 
\multicolumn{1}{|p{6.48cm}}{- Compute power: 10\textsuperscript{26} All models trained in excess of compute power} & 
\multicolumn{1}{|p{2.94cm}|}{\multirow{3}{*}{\parbox{2.94cm}{-.}}} \\ 
\hhline{~~-~}
\multicolumn{1}{|p{2.49cm}}{} & 
\multicolumn{1}{|p{5.74cm}}{} & 
\multicolumn{1}{|p{6.48cm}}{- Biological sequence data: All models trained primarily on biological sequence data 10\textsuperscript{23}} & 
\multicolumn{1}{|p{2.94cm}|}{} \\ 
\hhline{~~-~}
\multicolumn{1}{|p{2.49cm}}{} & 
\multicolumn{1}{|p{5.74cm}}{} & 
\multicolumn{1}{|p{6.48cm}}{- 10\textsuperscript{20} Cloud computing environments} & 
\multicolumn{1}{|p{2.94cm}|}{} \\ 
\hline
\multicolumn{1}{|p{2.49cm}}{Technical documentation} & 
\multicolumn{1}{|p{5.74cm}}{Vendors need to maintain up-to-date technical documentation, including training and testing details} & 
\multicolumn{1}{|p{6.48cm}}{Developers need to provide information on training, development, and physical and cybersecurity protections} & 
\multicolumn{1}{|p{2.94cm}|}{42001} \\ 
\hline
\multicolumn{1}{|p{2.49cm}}{Establish guidelines} & 
\multicolumn{1}{|p{5.74cm}}{Need to establish guidelines for copyright compliance} & 
\multicolumn{1}{|p{6.48cm}}{-.} & 
\multicolumn{1}{|p{2.94cm}|}{-.} \\ 
\hline
\multicolumn{1}{|p{2.49cm}}{Energy consumption} & 
\multicolumn{1}{|p{5.74cm}}{Need to report the model's estimated energy consumption} & 
\multicolumn{1}{|p{6.48cm}}{-.} & 
\multicolumn{1}{|p{2.94cm}|}{TR 20226} \\ 
\hline
\multicolumn{1}{|p{2.49cm}}{Security measures} & 
\multicolumn{1}{|p{5.74cm}}{Cybersecurity, physical safeguards} & 
\multicolumn{1}{|p{6.48cm}}{-.} & 
\multicolumn{1}{|p{2.94cm}|}{-.} \\ 
\hline
\multicolumn{1}{|p{2.49cm}}{Transparency} & 
\multicolumn{1}{|p{5.74cm}}{Logging features for tracing} & 
\multicolumn{1}{|p{6.48cm}}{-.} & 
\multicolumn{1}{|p{2.94cm}|}{24970} \\ 
\hline
\multicolumn{1}{|p{2.49cm}}{Transparency} & 
\multicolumn{1}{|p{5.74cm}}{Providers are required to publish learning data summaries} & 
\multicolumn{1}{|p{6.48cm}}{Need guidance on content attribution and labeling} & 
\multicolumn{1}{|p{2.94cm}|}{-.} \\ 
\hline
\multicolumn{1}{|p{2.49cm}}{Risk mitigation} & 
\multicolumn{1}{|p{5.74cm}}{Conduct comprehensive risk assessments, including real-world testing and serious incident reporting} & 
\multicolumn{1}{|p{6.48cm}}{AI red team testing and model guardrail development} & 
\multicolumn{1}{|p{2.94cm}|}{23894:2023} \\ 
\hline
\multicolumn{1}{|p{2.49cm}}{Compliance for high-risk AI} & 
\multicolumn{1}{|p{5.74cm}}{GPAI models used in high-risk scenarios require compliance with high-risk AI requirements} & 
\multicolumn{1}{|p{6.48cm}}{Dual-purpose foundational models are required to report ongoing/planned activities and AI red team test results} & 
\multicolumn{1}{|p{2.94cm}|}{TS 29119-11} \\ 
\hline
\multicolumn{1}{|p{2.49cm}}{\multirow{3}{*}{\parbox{2.49cm}{Systemic risk management}}} & 
\multicolumn{1}{|p{5.74cm}}{\multirow{3}{*}{\parbox{5.74cm}{GPAI providers are obligated to manage risk, conduct real-world testing, report serious incidents, and continuously monitor}}} & 
\multicolumn{1}{|p{6.48cm}}{\multirow{3}{*}{\parbox{6.48cm}{Need to implement guidelines and benchmarks for secure AI development and deployment}}} & 
\multicolumn{1}{|p{2.94cm}|}{42001} \\ 
\hhline{~~~-}
\multicolumn{1}{|p{2.49cm}}{} & 
\multicolumn{1}{|p{5.74cm}}{} & 
\multicolumn{1}{|p{6.48cm}}{} & 
\multicolumn{1}{|p{2.94cm}|}{23894:2023} \\ 
\hhline{~~~-}
\multicolumn{1}{|p{2.49cm}}{} & 
\multicolumn{1}{|p{5.74cm}}{} & 
\multicolumn{1}{|p{6.48cm}}{} & 
\multicolumn{1}{|p{2.94cm}|}{TS 29119-11} \\ 
\hline
\multicolumn{1}{|p{2.49cm}}{Privacy} & 
\multicolumn{1}{|p{5.74cm}}{Adhere to privacy, data governance, and transparency principles} & 
\multicolumn{1}{|p{6.48cm}}{Need to leverage privacy-enhancing technologies to protect your privacy} & 
\multicolumn{1}{|p{2.94cm}|}{-.} \\ 
\hline
\multicolumn{1}{|p{2.49cm}}{Conformance testing} & 
\multicolumn{1}{|p{5.74cm}}{Third-party conformance test results report} & 
\multicolumn{1}{|p{6.48cm}}{-.} & 
\multicolumn{1}{|p{2.94cm}|}{-.} \\ 
\hline
\end{tabular}
\end{adjustbox}
\caption{European/U.S. regulatory requirements and related standards}
\label{tab:europeanus_regulatory_requirements_and_related}\end{table}
\section{Conclusion}

This paper analyzed the trends in safety and trustworthiness standardization for advanced AI systems and considers the future direction of development. In the process, we find these kinds of considerable trends.

First, advanced AI technologies are evolving beyond traditional narrow domain AI to advanced AI systems such as large language models (LLMs) and general purpose artificial intelligence (AGI), which are demonstrating human-level performance in many areas, such as solving complex problems, processing sophisticated natural language, and performing multi-domain tasks.

Second, as advanced AI advances, there are growing concerns about its safety and trustworthiness. The increasing complexity and autonomy of advanced AI systems has the potential to create new risks, including uncontrollability, ethical decision-making, long-term socioeconomic impacts, and safety.

Third, international cooperation and standardization are essential to ensure advanced AI safety and trustworthiness, beyond the responses of individual companies or individual countries. To this end, we observed that various efforts are currently underway to apply consistent safety and trustworthiness standards and minimize potential risks through the development of international standards. It also confirmed that further international cooperation, including harmonization of global regulatory frameworks and standards, is needed for the common good of mankind.

Finally, it was identified that there is a need to classify new risks and develop new safety and trustworthiness assessment techniques and systems considering the complexity and autonomy of advanced AI systems. It was also identified that standardized test methods and techniques for the verification of advanced AI systems in various applications are needed, and research on the direction of alignment between regulations and standards is needed.

Securing safety and trustworthiness is an essential item in the research and development of advanced AI systems. As more and more research and development of advanced AI technologies will be conducted in the future, we hope that the nine trend frameworks presented in this article will help you understand the direction of cooperation between technologies and standards, establish a leading international standardization strategy, and plan a global response strategy.

\section*{Glossary}

\textbf{Safety} \ \ \ \ Freedom from unacceptable risk. Being able to control/manage risks and avoid or minimize those that do occur is how safety is achieved.

\textbf{Trustworthiness} \ \ \ \ The ability to meet stakeholder expectations in a verifiable manner. Stakeholder expectations include the following sub-characteristics: accountability, accuracy, availability, controllability, integrity, quality, reliability, resilience, robustness, safety, security, transparency, and usability.

\section*{Acronyms}

AGI\quad \quad \quad Artificial General Intelligence

ASI\quad \quad \quad Artificial Super Intelligence

FM\quad \quad \quad  Foundation Model

FMTI\quad \quad  Foundation Model Transparency Index

GPAI\quad \quad  General Purpose Artificial Intelligence

HAC\quad \quad \ Human-AI Collaboration

IEC\quad \quad \quad International Electrotechnical Commission

ISO\quad \quad \quad  International Organization for Standardization

JTC\quad \quad \quad  Joint Technical Committee

LLM\quad \quad \ \ Large Language Model

LLMOps\quad Large Language Model Operations

MLOps\quad \ \ Machine Learning Operations

NIST\quad \quad \ National Institute of Standards and Technology

RMF\quad \quad \ Risk Management Framework

SC\quad \quad \quad Sub Committee

SIL\quad \quad \ \ \ Safety Integrity Level

TCM\quad \quad Trustworthiness Characteristics Matrix

TR\quad \quad \quad Technical Report

TS\quad \quad \quad Technical Specification

V$\&$V\ \ \ \ \ \ \ \ Verification and Validation

\section*{Acknowledgments}
This work was supported by the Korea Medical Device Development Fund grant funded by the Korea government (the Ministry of Science and ICT, the Ministry of Trade, Industry and Energy, the Ministry of Health and Welfare, the Ministry of Food and Drug Safety)(Project Number: RS-2023-00208294)

\bibliographystyle{unsrt}  
\bibliography{main}

\begin{thebibliography}{10}

\bibitem{ukgovernment2023frontier}
Uk~Government.
\newblock Frontier ai: capabilities and risks.
\newblock \url{https://www.gov.uk/government/publications/frontier-ai-capabilities-and-risks-discussion-paper}, 2023.
\newblock discussion paper.

\bibitem{bengio2023managing}
Bengio Y and et~al.
\newblock Managing extreme ai risks amid rapid progress.
\newblock {\em Sci}, 384:842--845, 2023.

\bibitem{bommasani2022on}
Rishi Bommasani and et~al.
\newblock On the opportunities and risks of foundation models.
\newblock \url{https://arxiv.org/abs/2108.07258}, 2022.
\newblock arXiv preprint.

\bibitem{ukgovernmentfuture}
Uk~Government.
\newblock Future risks of frontier ai.
\newblock \url{https://assets.publishing.service.gov.uk/media/653bc393d10f3500139a6ac5/future-risks-of-frontier-ai-annex-a.pdf}, 2023.

\bibitem{ukgovernment2024}
Uk~Government.
\newblock International scientific report on the safety of advanced ai.
\newblock \url{https://www.gov.uk/government/publications/international-scientific-report-on-the-safety-of-advanced-ai}, 2024.

\bibitem{toner2022exploring}
Toner H and Acharya A.
\newblock Exploring clusters of research in three areas of ai safety.
\newblock \url{https://cset.georgetown.edu/publication/exploring-clusters-of-research-in-three-areas-of-ai-safety/}, 02 2022.

\bibitem{aisafety2023the}
Ai~Safety and Summit.
\newblock The bletchley declaration by countries attending the ai safety summit.
\newblock \url{https://www.gov.uk/government/publications/ai-safety-summit-2023-the-bletchley-declaration/the-bletchley-declaration-by-countries-attending-the-ai-safety-summit-1-2-november-2023}, 11 2023.

\bibitem{weidingertaxonomy}
Laura Weidinger and et~al.
\newblock Taxonomy of risks posed by language models.
\newblock {\em Proc. 2022 ACM Conf. Fairness, Accountability, Transparency}, 2022.

\bibitem{solaiman2024evaluating}
Irene Solaiman and et~al.
\newblock Evaluating the social impact of generative ai systems in systems and society.
\newblock \url{https://arxiv.org/abs/2306.05949}, 2024.
\newblock arXiv preprint.

\bibitem{wang2024aa}
Lei Wang and et~al.
\newblock A survey on large language model based autonomous agents.
\newblock \url{https://arxiv.org/abs/2308.11432}, 2024.
\newblock arXiv preprint.

\bibitem{kinniment2024evaluating}
Megan Kinniment and et~al.
\newblock Evaluating language-model agents on realistic autonomous tasks.
\newblock \url{https://arxiv.org/abs/2312.11671}, 2024.
\newblock arXiv preprint.

\bibitem{anderljung2023frontier}
Markus Anderljung and et~al.
\newblock Frontier ai regulation: Managing emerging risks to public safety.
\newblock \url{https://arxiv.org/abs/2307.03718}, 2023.
\newblock arXiv preprint.

\bibitem{un2024265}
Resolution A/RES/78/265.
\newblock Seizing the opportunities of safe, secure and trustworthy artificial intelligence systems for sustainable development, 2024.
\newblock UN General Assembly on 21 March 2024.

\bibitem{jonghongjeontcm}
Jonghong Jeon.
\newblock Trustworthiness characteristics matrix.
\newblock \url{https://github.com/hollobit/WG3_TCM}.
\newblock GitHub repository.

\bibitem{liu2024trustworthy}
Yang Liu and et~al.
\newblock Trustworthy llms: a survey and guideline for evaluating large language models' alignment.
\newblock \url{https://arxiv.org/abs/2308.05374}, 2024.
\newblock arXiv preprint.

\bibitem{sun2024}
Lichao Sun and et~al.
\newblock Trustllm: Trustworthiness in large language models.
\newblock \url{https://arxiv.org/abs/2401.05561}, 2024.
\newblock arXiv preprint.

\bibitem{stanfordthe}
Stanford Crfm.
\newblock The foundation model transparency index.
\newblock \url{https://crfm.stanford.edu/fmti/May-2024/index.html}.
\newblock Technical report.

\bibitem{bommasani2024}
Rishi Bommasani, Kevin Klyman, Shayne Longpre, Betty Xiong, Sayash Kapoor, Nestor Maslej, Arvind Narayanan, and Percy Liang.
\newblock Foundation model transparency reports.
\newblock \url{https://arxiv.org/pdf/2402.16268}, 2024.
\newblock arXiv preprint.

\bibitem{hendrycks2023an}
Dan Hendrycks and et~al.
\newblock An overview of catastrophic ai risks.
\newblock \url{https://arxiv.org/abs/2306.12001}, 2023.
\newblock arXiv preprint.

\bibitem{weidinger2024holistic}
Laura Weidinger and et~al.
\newblock Holistic safety and responsibility evaluations of advanced ai models.
\newblock \url{https://arxiv.org/abs/2404.14068}, 2024.
\newblock arXiv preprint.

\bibitem{nist}
NIST.
\newblock Artificial intelligence risk management framework (ai rmf 1.0).
\newblock \url{https://www.nist.gov/itl/ai-risk-management-framework}.
\newblock NIST AI 100-1.

\bibitem{abercrombie2024a}
Gavin Abercrombie and et~al.
\newblock A collaborative, human-centred taxonomy of ai, algorithmic, and automation harms.
\newblock \url{https://arxiv.org/pdf/2407.01294}, 2024.
\newblock arXiv preprint.

\bibitem{zeng2024ai}
Yi~Zeng and et~al.
\newblock Ai risk categorization decoded (air 2024): From government regulations to corporate policies.
\newblock \url{https://arxiv.org/pdf/2406.17864}, 2024.
\newblock arXiv preprint.

\bibitem{OECDterm}
OECD.
\newblock Defining ai incidents and related terms.
\newblock \url{https://www.oecd.org/en/publications/2024/05/defining-ai-incidents-and-related-terms_88d089ec.html}, 2024.
\newblock Working paper.

\bibitem{dong2024safeguarding}
Yi~Dong and et~al.
\newblock Safeguarding large language models: A survey.
\newblock \url{https://arxiv.org/abs/2406.02622}, 2024.
\newblock arXiv preprint.

\bibitem{xia2024}
Boming Xia and et~al.
\newblock An ai system evaluation framework for advancing ai safety: Terminology, taxonomy, lifecycle mapping.
\newblock \url{https://arxiv.org/abs/2404.05388}, 2024.
\newblock arXiv preprint.

\bibitem{morris2023levels}
Meredith~Ringel Morris and et~al.
\newblock Levels of agi for operationalizing progress on the path to agi.
\newblock \url{https://arxiv.org/pdf/2311.02462}, 2023.
\newblock arXiv preprint.

\bibitem{phuong2024evaluating}
Phuong M.
\newblock Evaluating frontier models for dangerous capabilities.
\newblock \url{https://arxiv.org/abs/2403.13793}, 2024.
\newblock arXiv preprint.

\bibitem{ji2022survey}
Ziwei Ji and et~al.
\newblock Survey of hallucination in natural language generation.
\newblock \url{https://arxiv.org/abs/2202.03629}, 2022.
\newblock arXiv preprint.

\bibitem{parkai}
Peter~S. Park and et~al.
\newblock Ai deception: A survey of examples, risks, and potential solutions.
\newblock {\em Patterns}, 5, 2023.

\bibitem{huang2023a}
Xiaowei Huang and et~al.
\newblock A survey of safety and trustworthiness of large language models through the lens of verification and validation.
\newblock \url{https://arxiv.org/abs/2305.11391}, 2023.
\newblock arXiv preprint.

\bibitem{minaee2024large}
Shervin Minaee and et~al.
\newblock Large language models: A survey.
\newblock \url{https://arxiv.org/pdf/2402.06196}, 2024.
\newblock arXiv preprint.

\bibitem{wang2023software}
Junjie Wang and et~al.
\newblock Software testing with large language models: Survey, landscape, and vision.
\newblock \url{https://arxiv.org/pdf/2307.07221}, 2023.
\newblock arXiv preprint.

\bibitem{hou2023large}
Xinyi Hou and et~al.
\newblock Large language models for software engineering: A systematic literature review.
\newblock \url{https://arxiv.org/abs/2308.10620}, 2023.
\newblock arXiv preprint.

\bibitem{chang2023a}
Yupeng Chang and et~al.
\newblock A survey on evaluation of large language models.
\newblock \url{https://arxiv.org/pdf/2307.03109}, 2023.
\newblock arXiv preprint.

\bibitem{guo2023evaluating}
Zishan Guo and et~al.
\newblock Evaluating large language models: A comprehensive survey.
\newblock \url{https://arxiv.org/pdf/2310.19736}, 2023.
\newblock arXiv preprint.

\bibitem{schulhoff2024}
Sander Schulhoff and et~al.
\newblock The prompt report: A systematic survey of prompting techniques.
\newblock \url{https://arxiv.org/pdf/2406.06608}, 2024.
\newblock arXiv preprint.

\bibitem{yue2024mmmu}
Xiang Yue and et~al.
\newblock Mmmu: A massive multi-discipline multimodal understanding and reasoning benchmark for expert agi.
\newblock \url{https://arxiv.org/abs/2311.16502}, 2024.
\newblock arXiv preprint.

\bibitem{zhu2024promptrobust}
Kaijie Zhu and et~al.
\newblock Promptrobust: Towards evaluating the robustness of large language models on adversarial prompts.
\newblock \url{https://arxiv.org/abs/2306.04528}, 2024.
\newblock arXiv preprint.

\bibitem{iec63521}
IEC~63521 ED1.
\newblock Machine learning-enabled medical device -performance evaluation process.
\newblock \url{https://www.iec.ch/dyn/www/f?p=103:38:411515690011080::::FSP_ORG_ID,FSP_APEX_PAGE,FSP_PROJECT_ID:1245,23,107066}.
\newblock IEC TC62 Committee Draft document.

\bibitem{wang2024ab}
Hong Wang and et~al.
\newblock A survey on an emerging safety challenge for autonomous vehicles: Safety of the intended functionality.
\newblock {\em Eng}, 33, 2024.

\bibitem{wang2023aligning}
Yufei Wang and et~al.
\newblock Aligning large language models with human: A survey.
\newblock \url{https://arxiv.org/abs/2307.12966}, 2023.
\newblock arXiv preprint.

\bibitem{shen2023large}
Tianhao Shen and et~al.
\newblock Large language model alignment: A survey.
\newblock \url{https://arxiv.org/abs/2309.15025}, 2023.
\newblock arXiv preprint.

\bibitem{ji2023}
Jiaming Ji and et~al.
\newblock Ai alignment: A comprehensive survey.
\newblock \url{https://arxiv.org/abs/2310.19852}, 2023.
\newblock arXiv preprint.

\bibitem{wang2024ac}
Zhichao Wang and et~al.
\newblock A comprehensive survey of llm alignment techniques.
\newblock \url{https://arxiv.org/abs/2407.16216}, 2024.
\newblock arXiv preprint.

\bibitem{vats2024a}
Vanshika Vats and et~al.
\newblock A survey on human-ai teaming with large pre-trained models.
\newblock \url{https://arxiv.org/abs/2403.04931}, 2024.
\newblock arXiv preprint.

\bibitem{univ2024}
Univ Stanford.
\newblock 2024 ai index report.
\newblock \url{https://aiindex.stanford.edu/report/}, 2024.

\bibitem{draganmintroducing}
Draganm A, King H, and Dafoe A.
\newblock Introducing the frontier safety framework.
\newblock \url{https://deepmind.google/discover/blog/introducing-the-frontier-safety-framework/}, 2024.
\newblock Deepmind.

\bibitem{ferdaustowards}
Md~Meftahul Ferdaus and et~al.
\newblock Towards trustworthy ai: A review of ethical and robust large language models.
\newblock \url{https://arxiv.org/abs/2407.13934}, 2024.
\newblock arXiv preprint.

\bibitem{kolt2024responsible}
Noam Kolt, Markus Anderljung, Joslyn Barnhart, Asher Brass, Kevin Esvelt, Gillian~K. Hadfield, Lennart Heim, Mikel Rodriguez, Jonas~B. Sandbrink, and Thomas Woodside.
\newblock Responsible reporting for frontier ai development.
\newblock \url{https://arxiv.org/abs/2404.02675}, 2024.
\newblock arXiv preprint.

\bibitem{díazrodríguez2023connecting}
N.D. Rodríguez and et~al.
\newblock Connecting the dots in trustworthy artificial intelligence: From ai principles, ethics, and key requirements to responsible ai systems and regulation.
\newblock {\em Inf. Fusion}, 99, 2023.

\bibitem{jung2023applicability}
Jung Kyuhwan, Jeon Jonghong, and Kim Hwiyoung.
\newblock Applicability and implications of llm in the medical field.
\newblock \url{https://www.maillink.co.kr/agency/html/20231221/REPORT1.pdf}, 2023.
\newblock Issue Report of the Korean Medical Information Society.

\bibitem{epoch2024notable}
Epoch AI.
\newblock Notable ai models.
\newblock \url{https://epochai.org/data/notable-ai-models}, 2024.
\newblock Technical Report.

\end{thebibliography}

\end{document}